\documentclass[10pt,twocolumn,letterpaper]{article}

\usepackage{cvpr}              % To produce the CAMERA-READY version
\usepackage{indentfirst}
\usepackage{float} 
\usepackage{overpic}
\usepackage{array}

\usepackage[dvipsnames]{xcolor}

\definecolor{cvprblue}{rgb}{0.21,0.49,0.74}
\usepackage[pagebackref,breaklinks,colorlinks,citecolor=cvprblue]{hyperref}

\def\paperID{7350} % *** Enter the Paper ID here
\def\confName{CVPR}
\def\confYear{2024}
\newcommand\nonumfootnote[1]{%
\begingroup%
    \renewcommand\thefootnote{}\footnote{\hspace{-3.7pt}#1}%
    \addtocounter{footnote}{-1}%
\endgroup%
}

\title{CosAvatar: Consistent and Animatable Portrait Video Tuning with Text Prompt}

\begin{document}
% \maketitle
\author{
    Haiyao Xiao\textsuperscript{1*} \quad
    Chenglai Zhong\textsuperscript{1*} \quad
    Xuan Gao\textsuperscript{1} \quad
    Yudong Guo\textsuperscript{2} \quad
    Juyong Zhang\textsuperscript{1$\dagger$} \\[5pt]
    $^1$University of Science and Technology of China \qquad
    $^2$Image Derivative Inc  \\[8pt]
    % \url{https://qiuyu96.github.io/CoDeF/}
}

\twocolumn[{
\maketitle

\vspace*{-9mm}

\begin{center}
   \begin{overpic}
        [width=\linewidth]{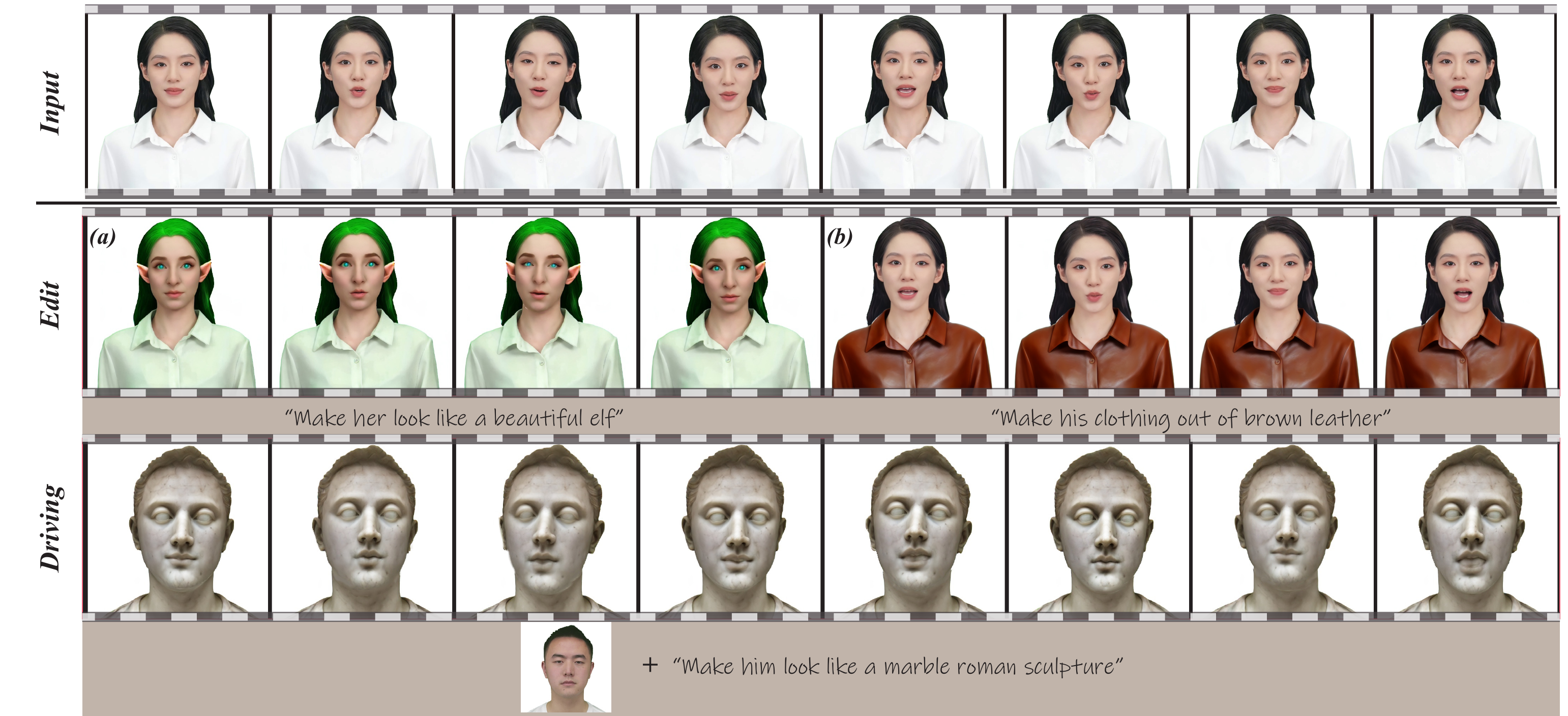}
        %  \fbox{\rule{0pt}{2in} \rule{linewidth}{0pt}}
   \end{overpic}
\end{center}
\vspace*{-5mm}
\captionof{figure}{CosAvatar, a text-driven portrait editing framework based on monocular dynamic NeRF. It allows for both (a) global style editing and (b) local attribute editing while ensuring strong consistency. It also enables expressive animation of the edited portrait.}

\label{fig:teaser}

\vspace*{5mm}

}]

% \begin{teaserfigure*}
%   \centering
%    \includegraphics[width=0.75\linewidth]{figures/teaser.pdf}
%    \caption{Teaser}
%    \label{fig:teaser}
% \end{teaserfigure*}
\nonumfootnote{* Equal contribution \qquad $\dagger$ Corresponding author}

\begin{abstract}
Recently, text-guided digital portrait editing has attracted more and more attentions. However, existing methods still struggle to maintain consistency across time, expression, and view or require specific data prerequisites. To solve these challenging problems, we propose CosAvatar, a high-quality and user-friendly framework for portrait tuning. With only monocular video and text instructions as input, we can produce animatable portraits with both temporal and 3D consistency. Different from methods that directly edit in the 2D domain, we employ a dynamic NeRF-based 3D portrait representation to model both the head and torso. We alternate between editing the video frames' dataset and updating the underlying 3D portrait until the edited frames reach 3D consistency. Additionally, we integrate the semantic portrait priors to enhance the edited results, allowing precise modifications in specified semantic areas. Extensive results demonstrate that our proposed method can not only accurately edit portrait styles or local attributes based on text instructions but also support expressive animation driven by a source video. Demo videos are provided in our project page: \href{https://ustc3dv.github.io/CosAvatar}{https://ustc3dv.github.io/CosAvatar}
% \nonumfootnote{* Equal contribution \qquad $\dagger$ Corresponding author}
\end{abstract}
\section{Introduction} \label{introduction}

Digital humans are widely used in AR/VR, gaming, film, and many other fields. Although the reconstruction and animation of digital humans have achieved remarkable success, there still remain unsolved problems in editing, especially cross-modality editing. How to edit the attributes of the portrait according to the given instruction while ensuring the structure similarity and its controllability is an important and challenging problem. Existing methods typically concentrate on editing static scenes or operate only within the image domain, resulting in a lack of temporal consistency and 3D consistency, and none of these methods could be animated by videos of other subjects. These constraints call for more flexible and generalizable methods for stylized portrait modeling.

% \gaoxuan{In recent years, researchers~\cite{yang2022Dual,yang2022Vtoonify} have been exploring the use of Generative Models for editing the appearance of portraits. These methods require a target image for style transfer on input portrait images or videos. However, compared to image inputs, text instructions offers a more convenient and intuitive approach for editing. The development of vision-language models, like CLIP~\cite{clip}, has effectively bridged the gap between vision and language, making it possible to edit and generate digital portrait based on textual input. Although the relevant CLIP-based methods~\cite{aneja2022clipface,wang2022clipnerf,kwon2022clipstyler,patashnik2021styleclip} have achieved promising results, in reality, their editing capabilities are limited.}

% \gaoxuan{Early works on portrait editing ~\cite{yang2022Dual,yang2022Vtoonify,liu20223dfm} often requires style labels or reference images to achieve the intended editing effect. However, compared to image inputs, text instructions offers a more user-friendly and intuitive approach for editing. The development of vision-language models, like CLIP~\cite{clip}, has effectively bridged the gap between vision and language, making it possible to edit and generate digital portrait based on textual input. Some works directly minimize the CLIP similarity loss function. This strategy is found lack of realism and accuracy. With the development of generative models like GAN or diffusion model, more and more works sought to use cross-modality generative models for portrait editing.}

Early works on portrait editing~\cite{yang2022Dual,yang2022Vtoonify,liu20223dfm} often require style labels or reference images to achieve the intended editing effect. However, compared with image inputs, text instructions offer a more user-friendly and intuitive approach to editing. The development of vision-language models, like CLIP~\cite{clip}, has effectively bridged the gap between vision and language, making it possible to edit and generate digital portraits based on textual input. Some work~\cite{aneja2023clipface,canfes2022textavatar,patashnik2021styleclip} directly minimizes the CLIP similarity loss function. This strategy is found to lack realism and accuracy. With the development of generative models like GAN and diffusion model, more and more works sought to use cross-modality generative models for portrait editing.

% In recent years, researchers~\cite{yang2022Dual,yang2022Vtoonify,liu20223dfm} have been exploring the use of Generative Models for editing the appearance of portraits. These methods require a target image as condition input to achieve the intended editing effect. However, compared to image inputs, text instructions offers a more user-friendly and intuitive approach for editing. The development of vision-language models, like CLIP~\cite{clip}, has effectively bridged the gap between vision and language, making it possible to edit and generate digital portrait based on textual input. Although the relevant CLIP-based methods~\cite{aneja2022clipface,kwon2022clipstyler,patashnik2021styleclip} have achieved promising results, in reality, their cross-modal generation capabilities are limited.

Recent GAN-based methods often adopt non-adversarial fine-tuning strategies under the guidance of CLIP similarity. As demonstrated in~\cite{kim2022datid3d}, these methods suffer from poor image quality and inferior text-image correspondence. With the tremendous success of diffusion models such as Imagen~\cite{saharia2022photorealistic}, DALL-E2~\cite{ramesh2022hierarchical}, and stable diffusion~\cite{rombach2022highresolution} in the domain of text-to-image generation, there has been a growing interest in leveraging these models to guide the generation or editing of digital human. Both 2D-based methods~\cite{brooks2022instructpix2pix,rombach2022highresolution,ruiz2023dreambooth} and 3D-based methods~\cite{instructnerf2023, poole2022dreamfusion} have achieved exciting results. However, these methods always focus on the construction of individual, static scenes, making it difficult for them to handle dynamic sequences directly. Some methods~\cite{khachatryan2023text2videozero,yang2023rerender,ouyang2023codef} utilize cross-frame consistency constraints or optical flow to enhance temporal consistency when dealing with dynamic scenes. However, due to the complexity of human motion in portrait videos, these techniques still generate obvious artifacts. In addition, some methods leverage 3D priors, such as explicit 3D parametric model~\cite{aneja2023clipface,zhang2023dreamface} or multi-view information~\cite{mendiratta2023avatarstudio} to enhance 3D consistency. However, the former suffers from the limited representation ability of the explicit parametric model, while the latter requires multi-view video sequences to train LDM~\cite{rombach2022highresolution} and portrait model, making it challenging to use in practice.

In this paper, we propose CosAvatar, which means cosplaying your digital avatar according to the given text instructions. With a monocular video sequence and a text instruction as input, CosAvatar can generate edited, animatable human portraits. Based on the given portrait video, we first finetune a generic parametric NeRF model similar to MetaHead~\cite{zhang2023metahead} to reconstruct a realistic digital portrait. The alias-free super-resolution module of MetaHead could eliminate the texture sticking of neural rendering, and the reconstructed 3D NeRF portrait could provide strong 3D priors and dynamic priors. Once the NeRF model is constructed, we use a structure-preserving image-conditioned diffusion model (InstructPix2Pix~\cite{brooks2022instructpix2pix}) to edit the rendered frames. Specifically, we alternate between editing the dataset of video frames and updating the underlying NeRF-based 3D portrait. This iterative approach continues until the edited frames achieve both 3D and temporal consistency.

Due to the inconsistent motion between the head and torso, we adopt the similar strategy used in AD-NeRF~\cite{guo2021adnerf} to model these two parts separately. To achieve more accurate editing of the local attributes in the portrait, we do not directly use the edited image to update the datasets but combine the edited image with the original image according to the semantic segmentation prior. The experimental results demonstrate that our method can generate structurally consistent and temporally coherent high-fidelity stylized portrait videos based on textual instructions. Additionally, it also allows precise control over certain semantic regions and is robust to arbitrary videos and complex expressions. It is also worth noting that our work requires only the information of the current frame for rendering. As a result, our system can handle video sequences of arbitrary length.
% \gaoxuan{It is also worthy to note that, because we only need the information of current frame for rendering, the demanded GPU memory of our CosAvatar doesn't increase with video length. Compared with works like TokenFlow~\cite{tokenflow2023} or CoDeF~\cite{ouyang2023codef} that are restricted to inference only 2-4s video sequence in NVIDIA 3090 GPU. Our work is able to inference a video sequence in arbitrary length.  }

In summary, the main contributions of this work include
\begin{itemize}
    \item We propose CosAvatar, a user-friendly portrait editing framework, which leverage parametric NeRF head representation and instruction-driven diffusion model to generate consistent and animatable Portrait Video 
    \item We leverage prior information from semantic segmentation to modify the fine-tuning process of our NeRF-based representation, allowing our method to perform editing on specified semantic regions.
    \item Our approach also enables animating the edited portrait based on subject from another video, while preserving both temporal and 3D coherence.
\end{itemize}
\section{Related Works}
% \haiyao{1. Text-Driven Diffusion Model(contain 2d 3d video, simplify), 2.Portrait Editing via Generative Models. 3.Portrait Modeling(HeadNeRF ,MetaHead)}

\noindent \textbf{Portrait Editing via Generative Models.} The editing of appearance and semantic attributes of digital humans has always attracted much attention. Following the success of StyleGAN2~\cite{karras2020styleGAN2} image generation, some reseachers~\cite{Abdal_2021_styleflow,liu20223dfm} extract facial attribute priors from virtual face rendering as condition input for styleGAN2, enabling the controlled editing of facial attributes. Other approaches~\cite{yang2022Dual,yang2022Vtoonify} focus on extracting attributes from given images and transferring them to the input image, which has also achieved promising results. The introduction of CLIP~\cite{clip} has established a bridge between vision and language, making it possible to edit and generate virtual avatars based on text instructions. These CLIP-based methods~\cite{kwon2022clipstyler,patashnik2021styleclip} allow for the customization of image generation to a wide range of styles from different domains based on the given text prompt. However, their cross-modal generation capabilities are limited and focus solely on editing and generation in the image domain, therefore lacking 3D consistency. Furthermore, its cross-modal generation capabilities are limited.

To address this issue, researchers utilize 3D representations as geometric proxies to enhance the 3D consistency cross editing. Some methods~\cite{sun2022fenerf,sun2023next3d} model the scene using NeRF and utilize generative models to fine-tune NeRF for editing purposes. These approaches can generate better quality and globally 3D consistent editing results. However, it often lacks direct drivability or the ability to perform local editing in specific regions. Other methods~\cite{canfes2022textavatar,aneja2023clipface} directly employ 3DMM (3D Morphable Model) as geometric representation and utilize generative models to generate corresponding UV textures, which enable avatar creation and driving for direct use in games or movies. However, the generation and editing capabilities of using a 3DMM are inadequate due to its limited representation ability.

\noindent \textbf{Text-Driven Diffusion Models.} Recent studies focus on diffusion models~\cite{ho2020denoising} for text-driven generation tasks, which can be classified into 2D image synthesis~\cite{brooks2022instructpix2pix,ruiz2023dreambooth,rombach2022highresolution,zhang2023adding} and 3D representation generation~\cite{cao2023dreamavatar,instructnerf2023,poole2022dreamfusion,huang2023dreamwaltz,zhang2023dreamface}. While these approaches can generate high-quality results from arbitrary text prompts, they mainly concentrate on generating or editing individual, static tasks and are not intended to directly manage dynamic scenes, especially portrait videos with complex motion.

As a result, some researchers have shifted their focus to video processing using text-to-image models. These methods could generate semantically corresponded video results directly based on text input~\cite{esser2023structure-gen1,singer2022makeavideo,ho2022imagen}, but precise control over the results remains a challenge. To solve this problem, some researchers~\cite{wu2023tuneavideo,qi2023fatezero,wang2023zeroshot} modify the latent space of the Diffusion model and introduce cross-frame attention maps to enhance the consistency of the generated results. However, the consistency in textures and details remains unsatisfactory. Rerender-A-Video~\cite{yang2023rerender} proposes a hierarchical cross-frame consistency constraint to achieve global and local temporal consistency. CoDeF~\cite{ouyang2023codef} learns a canonical content field and a temporal deformation field to represent the video. Simply editing the canonical images will naturally propagate to the entire video with the aid of the temporal deformation field. As a result, it also achieves superior cross-frame consistency. However, both methods require optical flow information as additional input, which limits their performance in handling complex motion and long videos. Avatarstudio~\cite{mendiratta2023avatarstudio} follow the avatar modeling approach presented in HQ3DAvatar~\cite{teotia2023hq3davatar} and proposal a view-and-time-aware Score Distillation Sampling to enable high-quality personalized editing across the view and time domain. However, it requires multi-view dynamic video sequences as input for avatar modeling, which can be difficult to obtain for practical use.

\noindent \textbf{Parametric Head Model.} The parametric head model embeds the human head into low-dimensional identity, expression, and appearance space. Blanz and Vetter proposed 3DMM~\cite{blanz1999morphable} to embed 3D head shape into several low-dimensional PCA spaces. 
Mesh-based parametric head model has been further studied by a lot of works. Multilinear models~\cite{cao2013facewarehouse,vlasic2006face}, non-linear models~\cite{ranjan2018generating,tran2018nonlinear}, articulated models~\cite{li2017learning} are used to improve the representation ability.

\begin{figure*}[t]
  \centering
   \includegraphics[width=\linewidth]{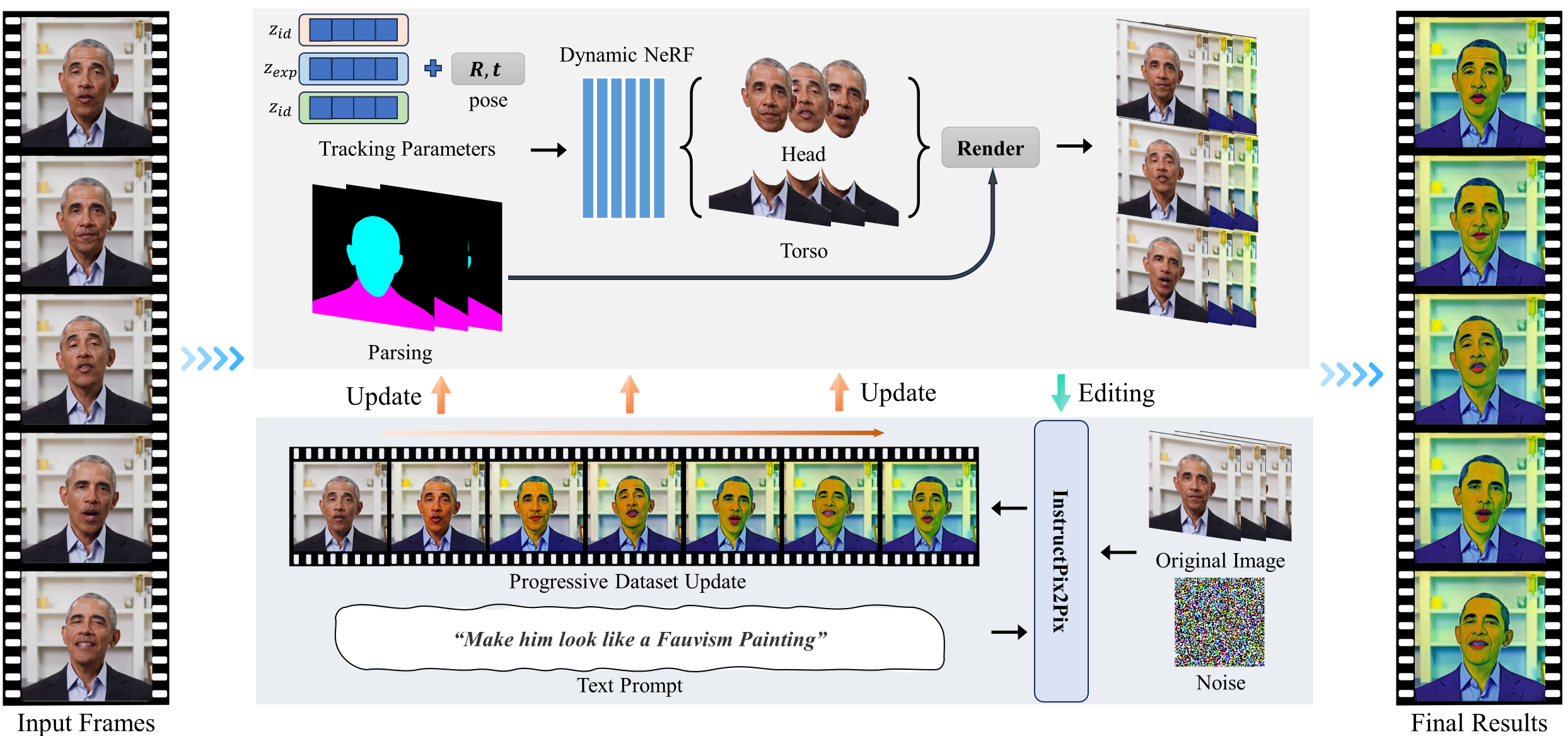}
   \caption{Algorithm pipeline of CosAvatar. We train two dynamic neural radiance fields to reconstruct the head and torso region separately based on the tracking coefficient and pose. Once the NeRF model is constructed, we use Instruct-Pix2Pix to generate edited results based on the instruction and progressively update the sequence datasets and NeRF model. In the end, through the utilization of the NeRF-based portrait representation, we are able to produce an editing sequence that maintains temporary coherence and 3D consistency.}
   \label{fig:pipe}
\end{figure*}

Recently, implicit head models have been discovered to represent digital humans. Compared with mesh-based models, implicit head models have higher representation ability and are differentiable in rendering. There have been extensive works on subject specific head modeling~\cite{Gafni_2021_CVPR,zielonka2022instant,xu2023avatarmav,zheng2022imface,zheng2022avatar}. Generic parametric head models~\cite{yenamandra2021i3dmm,hong2021headnerf,zhuang2022mofanerf,zhang2023metahead} turn to disentangle the latent spaces of human heads. HeadNeRF~\cite{hong2021headnerf} disentangles identity, expression, and appearance spaces. MetaHead~\cite{zhang2023metahead} presents a unified and full-featured controllable digital head engine. There are also works focus on efficient training/inference~\cite{Gao2022nerfblendshape,zielonka2022instant,xu2023avatarmav}, cross-modality driving~\cite{guo2021adnerf,liu2022semantic,shen2022dfrf} and generation~\cite{chan2022efficient,wang2023rodin,deng2022gram,sun2023next3d}.

\section{Methods}
Our goal is to generate stylized dynamic human portrait sequences comprising both the head and torso regions by training on a given monocular dynamic portrait video and natural language editing instructions as additional input. Given a portrait video, we first leverage the existing methods to estimate semantic segmentation~\cite{yu2018bisenet}. Then, we solve an inverse rendering optimization based on the FLAME\cite{li2017learning} model to get the corresponding semantic parameters per frame. Based on the preprocessed data, we represent human portraits through a NeRF-based dynamic model, where the dynamic head and torso are modeled separately. The deformation of the portrait is conditioned by the estimated FLAME parameters (Sec.~\ref{method32}). Then, we leverage existing large text-guided latent diffusion models to generate stylized images based on natural language input and semantic segmentation results. These generated images are then used to update the previous dynamic portrait model (Sec.~\ref{method33}). Experiments illustrate that our method can generate high-fidelity editing results for human portraits that align with the provided editing instructions.

\subsection{Dynamic Neural Radiance Fields for Potrait} 
\label{method32}

% \haiyao{TODO: Add description for MetaHead}
As previously mentioned, our method utilizes a NeRF architecture similar to MetaHead~\cite{zhang2023metahead} and incorporates a deformation field to capture portrait motion. This design enables us to perform fine-tuning directly on top of the MetaHead, accelerating the reconstruction process.
Since the movement of the head part is not always consistent with the torso, we opt to model them separately. One model is specifically designed to capture the dynamic behavior of the head, while another model focuses on the torso region. Both of these utilize the same NeRF architecture but employ different control signals.

Specifically, given a frame in the input 2D sequence, first we sample the points along the randomly casted camera rays $\mathbf{d}$ in the world space, denoted as $\mathbf{x} \in \mathbf{R}^3$. A deformation field $\mathcal{D}_\theta$ is used to model the dynamic variations of portrait, which takes as input a point in world space and deformation latent vector $\mathbf{w}$, and regresses a deformation field that converts the world point $\mathbf{x}$ to a canonical space $\hat{\mathbf{x}}$, as follows:
\begin{equation}
    \hat{\mathbf{x}} = \mathbf{x} + \mathcal{D}_\theta(\gamma(\mathbf{x}), \mathbf{w}),
\end{equation}
where $\gamma$ represents the positional encoding used in~\cite{mildenhall2020nerf}. For the head part, since we have converted the head pose into the camera parameters, we believe that the deformation of the head should be independently controlled by the expression. Therefore, we set $\mathbf{w}$ as the FLAME expression parameters $\mathbf{z}_{\textrm{exp}}$ estimated of each frame during the data preprocessing stage and keep them fixed. For the torso part, since we only use the upper body part after cropping and do not contain the arms, we believe that it should mainly consist of rigid deformation. Therefore, we randomly initialize the deformation latent code $\mathbf{w}$ of all frames to the same value and optimize them together with the network. To better leverage the generalization ability of MetaHead for fast reconstruction, we retain the identity and lighting parameters as inputs to the NeRF model.

After that, we learn the conditional radiance field in the canonical space using implicit function $\mathcal{F}_\theta$ as follows:
\begin{equation}
    \mathcal{F}_\theta:  (\gamma(\hat{\mathbf{x}}), \gamma(d), \mathbf{z}_{\textrm{id}}, \mathbf{z}_{\textrm{exp}},\mathbf{z}_{\textrm{ill}}) \rightarrow (\sigma, \mathbf{F})
\end{equation}

Like previous works~\cite{hong2021headnerf, zhang2023metahead}, instead of directly predicting RGB color, we predict a high-dimensional feature vector $\mathbf{F}(\mathbf{x}) \in \mathcal{R}^{256}$ for the 3D sampling point $\mathbf{x}$. Specifically, $\mathcal{F}_\theta$ takes as input the concatenation of $(\gamma(x), \mathbf{z}_{\textrm{id}})$ and output the density $\sigma$ of $\mathbf{x}$ and an intermediate feature, the latter and $(\gamma(d),\mathbf{z}_{\textrm{exp}},\mathbf{z}_{\textrm{ill}})$ are used to further predict $\mathbf{F}(\mathbf{x})$. After that, the final pixel color of feature map $\mathbf{I}_{F}$ is given by volume rendering:
\begin{equation}
\begin{aligned}
I_{F}(d)&=\int_{0}^{\infty} w(t) \cdot F(d(t)) dt \\
\textrm{where}  \quad w(t)&=\exp \left(-\int_{0}^{t} \sigma(d(s)) d s\right) \cdot \sigma(d(t)).
\end{aligned}
\end{equation}
% Finally, we add a super-resolution module $\mathbf{\theta}_{\textrm{sup}}$ used in ~\cite{zhang2023metahead} to improved the visual quality of the output portraits. It design a hierarchical structural attention module customized for 3D portraits vision tasks in each synthesis block of $\mathbf{\theta}_{\textrm{sup}}$, which can efficiently eliminate the structure distortion and 3D texture flickering. Like ~\cite{guo2021adnerf}, we will train two implicit model $\mathcal{F}^{\textrm{head}}_\theta$ and $\mathcal{F}^{\textrm{torso}}_\theta$ to generate images of the head and torso respectively, and apply two super-resolution modules $\mathcal{\theta}^{\textrm{head}}_{\textrm{sup}}$, $\mathcal{\theta}^{\textrm{torso}}_{\textrm{sup}}$ to process them. Then we combine them according to the parsing results to obtain the final high-resolution result, i.e

Finally, we add a super-resolution module $U$ used in~\cite{zhang2023metahead} to improve the visual quality of the output portraits. It design a hierarchical structural attention module customized for 3D portrait vision tasks in each synthesis block of $U$, which can efficiently eliminate the structure distortion and 3D texture flickering. Like~\cite{guo2021adnerf}, we will train two implicit models $\mathcal{F}^{\textrm{head}}_\theta$ and $\mathcal{F}^{\textrm{torso}}_\theta$ to generate images of the head and torso respectively, and apply two super-resolution modules $U^{\textrm{head}}$, $U^{\textrm{torso}}$ to process them. Then, we combine them according to the parsing results to obtain the final high-resolution result, i.e
% \begin{equation}
%     I_r = S_{\textrm{head}} \odot \mathbf{\theta}^{\textrm{head}}_{\textrm{sup}}(\mathcal{F}^{\textrm{head}}_\phi) + S_{\textrm{torso}} \odot \mathbf{\theta}^{\textrm{torso}}_{\textrm{sup}}(\mathcal{F}^{\textrm{torso}}_\phi)
% \end{equation}

\begin{equation}
    I_r = S_{\textrm{head}} \odot U^{\textrm{head}}(\mathcal{F}^{\textrm{head}}_\theta) + S_{\textrm{torso}} \odot U^{\textrm{torso}}(\mathcal{F}^{\textrm{torso}}_\theta)
\end{equation}

% \haiyao{TODO:Modify formula}
where $S_{\textrm{head}}$ and $S_{\textrm{torso}}$ means the head part and torso part obtained from the results of the face parsing~\cite{yu2018bisenet} applied to the input image.
\par

\subsection{Text-guided Dynamic NeRF Editing} \label{method33}
% \haiyao{TODO: Add specific description of Editing Model and Loss Function, divided into two subsections }

\noindent \textbf{Portrait editing based on the diffusion model.}
In order to edit dynamic 3D human portraits using a text prompt, we leverage Latent diffusion models (LDMs)\cite{rombach2022highresolution} to optimize the NeRF-based network. However, simply using the original diffusion model would soon lead to characteristic drift, i.e., properties such as identity, expression details, and motion of the portrait in the video are difficult to maintain. To avoid this, we adopt an approach similar to the one described in\cite{instructnerf2023}, where 2D images will be iteratively edited using Instruct-Pix2Pix~\cite{brooks2022instructpix2pix} and the dynamic 3D NeRF model will be iteratively optimized. By doing this, we can achieve 3D-consistent edits and preserve global style features. However, it may not offer precise control of local attributes, such as changing individual hair color or clothes. To improve our ability for local attribute editing, we are exploring adjustments based on the design mentioned above.

To provide more specific details, we also use InstructPix2Pix~\cite{brooks2022instructpix2pix} to edit each image in the video. It takes three inputs, an input RGB image $\mathbf{I}\in \mathcal{R}^{H\times W \times3}$, a text instruction $\mathbf{T}$, and a noisy latent code $\mathbf{z}_t$. We set $\mathbf{I}$ to the original image $\mathbf{I}_{gt}$ in the dataset. For $\mathbf{z}_t$, we set it a linear combination of $\mathcal{N}(0,1)$ and all latent images created by encoding current NeRF rendering results $\mathbf{z}_0 = \mathcal{E}(I_r)$. The diffusion model predicts the amount of noise present in the input $\mathbf{z}_t$, using the denoising U-Net $\epsilon_\theta$ as:
\begin{equation}
    \hat{\epsilon} = \epsilon_\theta(\mathbf{z}_t, t,\mathbf{I}, \mathbf{T}),
\end{equation}
where noise level $t$ is chosen at random from a constant range $[t_{\textrm{min}}, t_{\textrm{max}}]$. The noise prediction $\hat{\epsilon}$ then be used to derive output latent code $\hat{\mathbf{z}}$ via the DDIM sampling process and use a decoder to reconstruct the estimated image $\hat{\mathbf{I}}$ from the output latent code, i.e. $\hat{\mathbf{I}} = \mathcal{D}(\hat{\mathbf{z}})$. As the diffusion model always takes the original frame as input, it can prevent characteristic drift and keep coarse-grained image features unchanged during generation.

Building on this basis, InstructPix2Pix adopts Classifier-Free Diffusion Guidance~\cite{ho2022classifier} to get a modified score to compose both image condition $I$ and text condition $T$. The score used for denoising could be written as:
% \gaoxuan{InstructPix2Pix adopt Classifier-Free Diffusion Guidance~\cite{ho2022classifier} to get a modified score to compose both image condition $I$ and text condition $T$. The score used for denoising could be written as:}

% \vspace{-3.5mm}
\begin{equation}
\begin{split}
    \hat{\epsilon}_{\theta}(z_t, I, T) = &\: \epsilon_{\theta}(z_t, \varnothing, \varnothing) \\ &+ s_I \cdot (\epsilon_{\theta}(z_t, I, \varnothing) - \epsilon_{\theta}(z_t, \varnothing, \varnothing)) \\ &+ s_T \cdot (\epsilon_{\theta}(z_t, I, T) - \epsilon_{\theta}(z_t, I, \varnothing))
    \label{eq:cfg}
\end{split}
\end{equation}

The guidance scale $s_I$ effectively shifts probability mass toward data where an implicit classifier $p_{\theta}(I|z_t)$ assigns high likelihood to the image conditioning $I$, and guidance scale $s_T$ effectively shifts probability mass toward data where an implicit classifier $p_{\theta}(T|I,z_t)$ assigns high likelihood to the text instruction conditioning $T$. The balance between $s_I$ and $s_T$ trades off how strongly the generated samples correspond with the input image and how strongly they correspond with the edit instruction.
% \gaoxuan{Guidance scale $s_I$ effectively shifts probability mass toward data where an implicit classifier $p_{\theta}(I|z_t)$ assigns high likelihood to the image conditioning $I$, and guidance scale $s_T$ effectively shifts probability mass toward data where an implicit classifier $p_{\theta}(T|I,z_t)$ assigns high likelihood to the text instruction conditioning $T$. The balance between $s_I$ and $s_T$ trades off how strongly the generated samples correspond with the input image and how strongly they correspond with the edit instruction.}
\begin{figure}[h]
  \centering
   \includegraphics[width=1.0\linewidth]{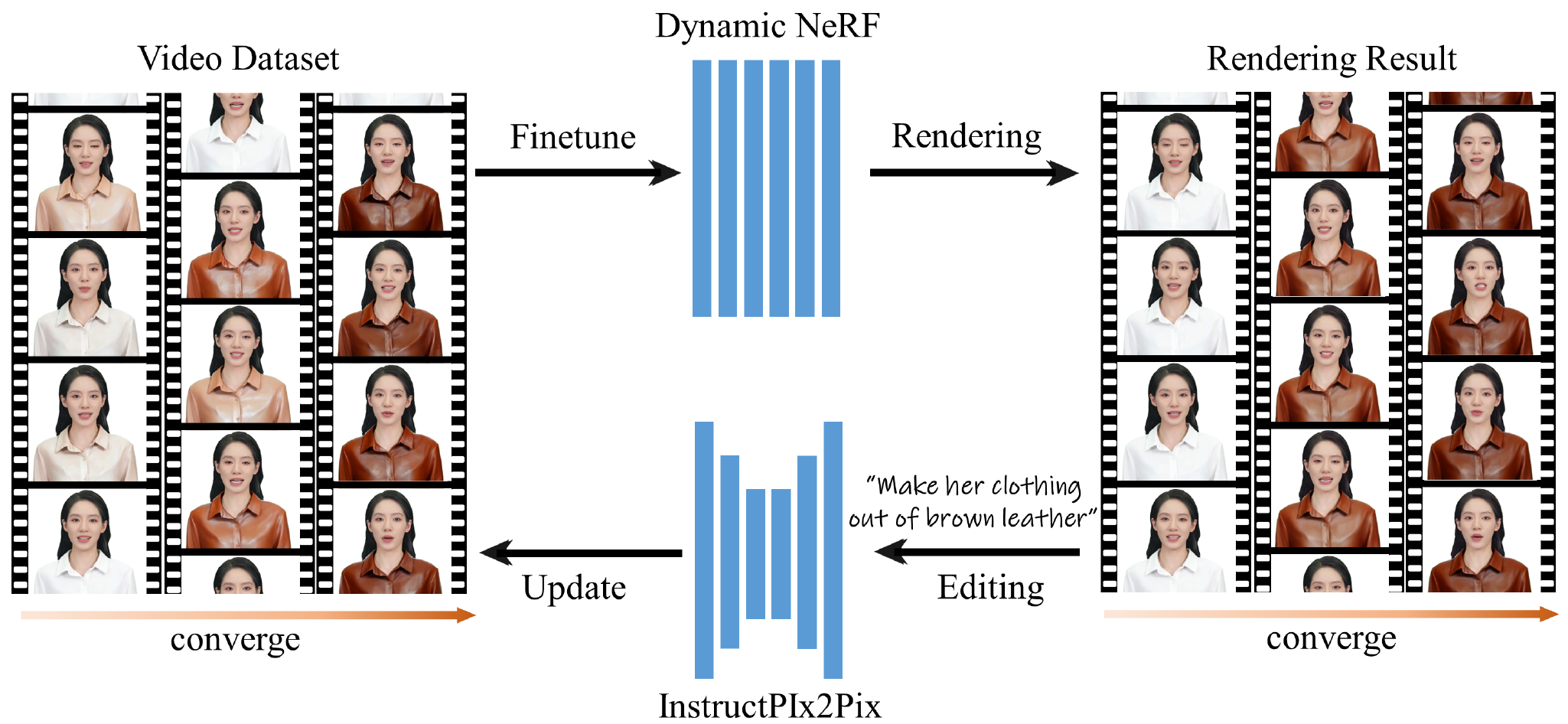}
   \caption{Explanation of the training strategy for CosAvatar.}
   \label{fig:DatasetUpdate}
\end{figure}

\begin{figure*}[t]
  \centering
   \includegraphics[width=\linewidth]{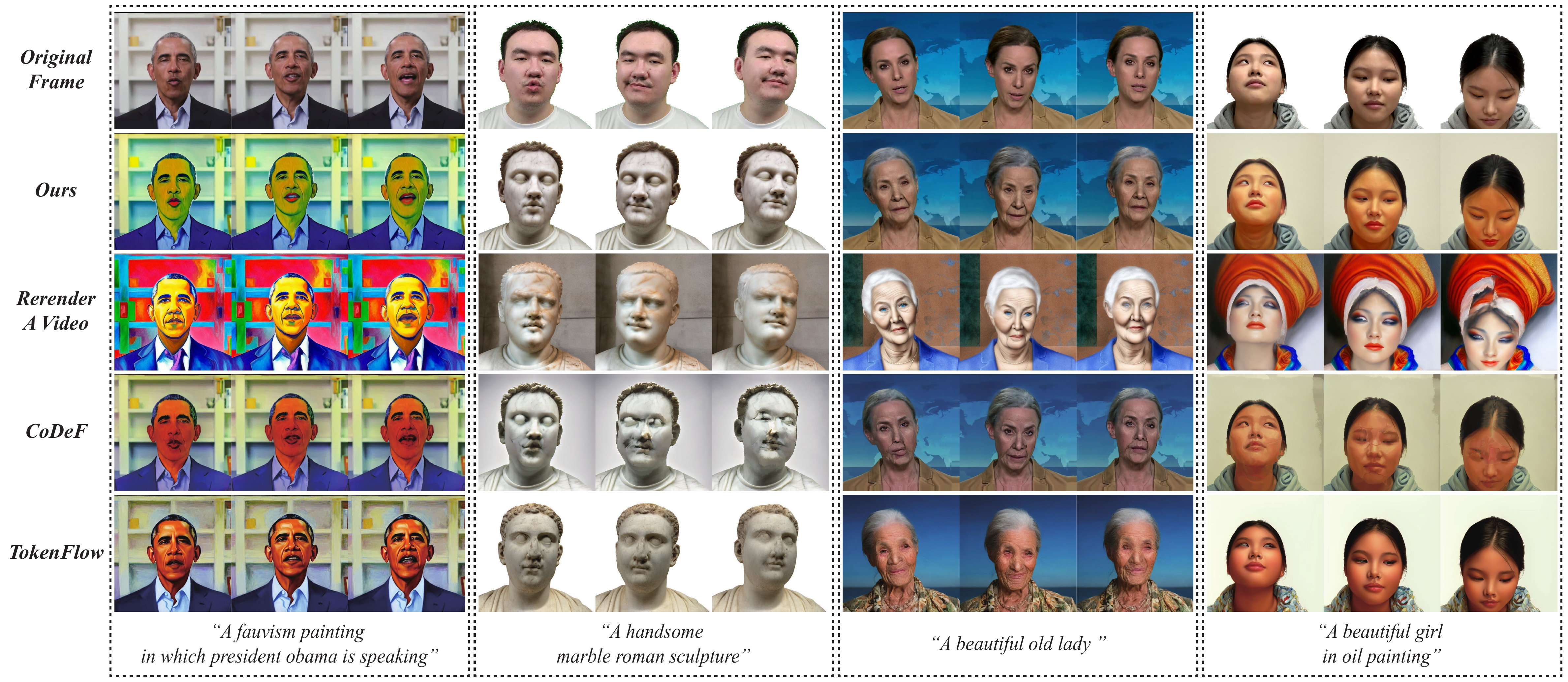}
   \caption{Qualitative comparison of global style editing with \textit{Rerender-A-Video}~\cite{yang2023rerender}, \textit{CoDeF}~\cite{ouyang2023codef}, and \textit{TokenFlow}~\cite{tokenflow2023}. Although these methods effectively preserve stylistic consistency on a global scale, artifacts still exist when it comes to portrait motion.}
   \label{fig:exp_total}
   \vspace*{-2mm}
\end{figure*}

\noindent \textbf{Progressive Dataset update.}
% \textcolor{red}{TODO}: dataset update according to parsing
We follow the basic strategy of dataset update in~\cite{instructnerf2023}. After NeRF is updated a certain number of times, we use the diffusion model to generate a new edited image $\mathbf{I}_e^0$ based on the current rendering result. Subsequently, we replace the corresponding images within the dataset with this newly generated result. In other words, the dataset will be semi-permanent updated during the fine-tuning process. As images are used to update the NeRF and progressively updated, they will gradually converge to achieve both 3D and temporal consistency together with NeRF. This convergence process is visually demonstrated in Fig.~\ref{fig:DatasetUpdate}. However, to edit the semantic area of the portrait more accurately, we do not directly use the edited image but combine it with the original image according to the parsing results. Specifically, we first retrieve the keywords in the text instruction and determine whether to edit the local semantic area. Then, we find the corresponding semantic area $S_{key}$ based on the parsing result of the original image. The final edit image is
\begin{equation}
    \mathbf{I}_e = S_{\textrm{key}} \odot \mathbf{I}_e^0+(1-S_{\textrm{key}}) \odot \mathbf{I}_{\textrm{gt}}
\end{equation}

To perform text-driven editing, we update only the NeRF function $\mathcal{F}_{\theta}$ and the super-resolution module $U$ while keeping the deformation prediction network $\mathcal{D}_\theta$ fixed. This is based on the assumption that the editing process should primarily modify the appearance within the canonical space and not impact the motion information in the video.

\subsection{Training Details}
% \haiyao{Will keep the Data Preprocess but will simplify the description}
\noindent \textbf{Datasets Setting.} Our method is trained using monocular portrait talking videos, which can be captured by a smartphone. We use the smartphone's front camera to capture and turn off the AutoFocus to prevent blurring. The recorded videos encompass the entire head and upper body above the chest region. They are captured at a resolution of 2K with a frame rate of 25 frames per second. On average, the duration of these videos was approximately 2 minutes. We assume that the recording camera and the background remain static throughout the recording process. In addition to our own captured videos, we also incorporate videos collected from NeRFBlendshape~\cite{Gao2022nerfblendshape} to evaluate the performance of our approach since their videos contain exaggerated expressions and large head rotations.\par

\noindent \textbf{Data Preprocessing.} The dataset preprocessing involves two main steps. Firstly, we perform image cropping and resizing to achieve a resolution of $512 \times 512$. Additionally, we utilize an automatic parsing method\cite{yu2018bisenet} to label the different semantic regions within each frame.
 
Then, we solve an inverse rendering optimization based on the FLAME model~\cite{yu2018bisenet}. We optimize the camera parameters and pre-frame FLAME coefficient by minimizing the distance between the 3D landmarks projection results and the ground truth landmark and the photometric error between rendering results and the GT image. Like HeadNeRF~\cite{hong2021headnerf}, we take the human head pose parameters as the extrinsic camera parameter of the corresponding frame, which implicitly aligns the underlying geometry of each frame to the same spatial location. 

\noindent \textbf{Loss Functions.} We optimize the radiance field using two main constraints. The first one is photo-metric reconstruction loss between the rendered image $\mathbf{I}_r$ and the editing results $\mathbf{I}_e$ stored in current datasets. The loss function is represented by: 
\begin{equation}
    \mathcal{L}_{\textrm{photo}} = \left\|\mathbf{I}_{\mathrm{r}} - \mathbf{I}_{\mathrm{e}} \right\|^{2}
\end{equation}

Second, we add a perceptual loss to further improve the image details, which calculate the feature distance between the rendered image and the editing results in a pre-trained VGG network. The loss function is formulated as follows:
\begin{equation}
    \mathcal{L}_{\textrm{perp}} = \sum_{l} \lambda_{l} \left\|\phi_{i}\left(\mathbf{I}_{\mathrm{r}}\right)-\phi_{i}\left(\mathbf{I}_{\mathrm{e}}\right)\right\|^{2},
\end{equation}
where $\lambda_{l}(*)$ denotes the activation of the l-th layer in VGG16 network. In summary, the overall loss of our methods is defined as:
\begin{equation} \label{eq10}
    \mathcal{L}=\mathcal{L}_{\textrm{photo}}+\alpha\mathcal{L}_{\textrm{perp}}.
\end{equation}
\section{Experiments}

\begin{figure*}[t]
  \centering
   \includegraphics[width=\linewidth]{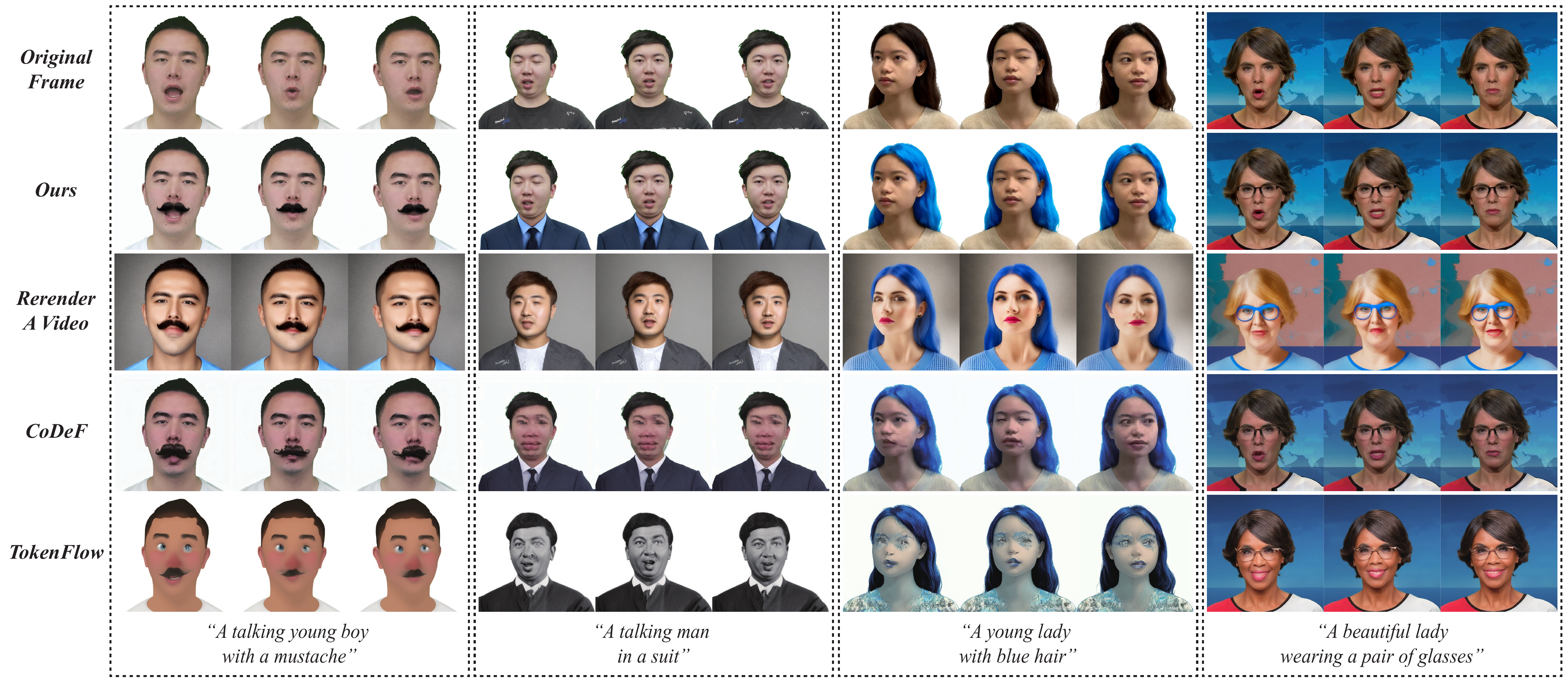}
   \caption{Qualitative comparison of local attribute editing with \textit{Rerender-A-Video}~\cite{yang2023rerender}, \textit{CoDeF}~\cite{ouyang2023codef}, and \textit{TokenFlow}~\cite{tokenflow2023}. }
   \label{fig:exp_region}
   \vspace{-2mm}
\end{figure*}
% \haiyao{Add some dataset description?}

% \subsection{Data Capture} \label{method31}
% \gaoxuan{I think it is not necessary to descbribe face tracking details, maybe we could move this part into supplemental materials or appendix? Will it be better to move capture settings into experiments section?}

% This enables us to utilize generalization models, such as MetaHead~\cite{zhang2023metahead}, for rapid reconstruction of new portrait video.
\subsection{Implementation details}
We implement our framework in PyTorch~\cite{pytorch}, and both networks are trained with Adam~\cite{kingma2017adam} solver with an initial learning rate of 0.0001 on a single NVIDIA 3090 GPU. For the dynamic portrait modeling, we will utilize the FLAME fitting results obtained during the data processing to compute a depth map and use it to guide the sampling. To further speed up training, We sampled 64 points along each ray and removed the hierarchical volume sampling of NeRF. This stage will take one hour for a video with 2000 frames for the best performance. As for the portrait editing, to accelerate the editing process, we set the text guidance scale $s_T$ to 12 and image guidance scale $s_I$ to 1.5 and the value of noise level range $t_{min}$ and $t_{max}$ is set to 0.25 and 0.95. In addition, We update one image after optimizing the NeRF model for ten iterations and perform 25 denoising steps for each update. Finally, we train our model for 50k iterations for the given text instruction, which takes approximately 4 hours on a single NVIDIA 3090 GPU. The loss weights in Eq. (\ref{eq10}) are set to $\alpha=0.5$.

\subsection{Evaluations}
% \gaoxuan{We compare CosAvatar with state-of-the-art portrait video editing methods, including TokenFlow~\cite{tokenflow2023}, Rerender-A-Video~\cite{yang2023rerender}, and CoDeF~\cite{ouyang2023codef}, for portrait video editing tasks. Because the demanded GPU memory of both CoDeF and TokenFlow increased with video length, they are not suitable for processing long video sequence. Therefore, we only select a 2-4 seconds' segment from every video, approximately 50-100 frames. For Rerender-A-Video and TokenFlow, we applied the same text instructions to edit the input videos. For CoDeF, we followed their provided workflow, which involved training the deformation field and canonical image on the original video first. Then, we used Instruct-Pix2Pix to edit the canonical image and generate the final edited video according to the deformation field. }
We compare CosAvatar with state-of-the-art video editing methods, including TokenFlow~\cite{tokenflow2023}, Rerender-A-Video~\cite{yang2023rerender}, and CoDeF~\cite{ouyang2023codef}, for portrait video editing tasks. 
For Rerender-A-Video and TokenFlow, we apply the same text instructions to edit the videos. For CoDeF, we follow their provided workflow, which involves training the deformation field and canonical image on the input video first. Then, we use Instruct-Pix2Pix to edit the canonical image and generate the final edited video according to the deformation field. 
To ensure a fair comparison, we limit the video segments used in our evaluation to 2-4 seconds, consisting of approximately 50-100 frames. This is necessary because TokenFlow's GPU memory requirements increase with the length of the video, and CoDeF's approach of learning a single canonical image and deformation field for the entire video sequence is not suitable for processing long videos. Although our method can handle videos of arbitrary length, selecting shorter segments allows for consistent evaluation and unbiased comparisons between different methods.
As shown in Fig.~\ref{fig:exp_total}, we evaluate the temporal consistency of the edited outcomes. Although these methods can generate edited results with consistent styles, they still struggle to maintain the consistency of fine-grained information. We found that sometimes the expressions of TokenFlow's and Rerender-A-Video's results are not similar to the original video, and artifacts may occur in the region of the eyes and mouth. This discrepancy may be attributed to the limitations of extended attention in ensuring detailed consistency, particularly in capturing facial expressions. Incorrect correspondences of NN search or optical flow estimation further amplify the inaccuracy. Although CoDeF's unique modeling approach improves its ability to preserve fine-grained consistency in short videos, it fails to generate reasonable results when faced with exaggerated expression and pose changes. This is because the 2D deformation field struggles to model complex 3D portrait deformations.

In contrast, our approach leverages NeRF as the geometric representation, providing superior 3D consistency. By incorporating prior information about the head as guidance, our method accurately captures motion in expression and posture, thereby ensuring temporal consistency in the edited results. Additionally, as illustrated in Fig.~\ref{fig:exp_region}, our method excels in local attribute editing by introducing semantic segmentation. Our method effectively preserves other semantic attributes throughout this process, particularly the identity of the edited portrait, which is a challenging task for other methods. A comprehensive comparison can be better achieved by watching the accompanying video.

For quantitative evaluation, we adopt the metrics `Clip-Text', `Tem-Con', and `Pixel-MSE' from Rerender-a-Video~\cite{yang2023rerender} and Pix2Video~\cite{ceylan2023pix2video}, as presented in Tab.~\ref{tab:quantitative}. `Clip-Text' calculates the cosine similarity between the CLIP embedding of the edit prompt and the embedding of each frame in the edited video. `Tem-Con' and `Pixel-MSE' assess the consistency by measuring the CLIP-based cosine similarity and the averaged mean squared pixel error between consecutive frame pairs. The results indicate that our method outperforms others in terms of temporal consistency.

\begin{table}
  \centering
  \renewcommand{\arraystretch}{1.15}
  \begin{tabular}{@{}l|c|c|c@{}}
    \toprule
    Metrics & Clip-Text$\uparrow$ & Tem-Con$\uparrow$ & Pixel-MSE$\downarrow$  \\
    \midrule
    CoDeF & 0.2979 & 0.979 & 0.051  \\
    TokenFlow & 0.3064 & 0.981 & 0.059  \\
    RAV & \textbf{0.3123} & 0.984 & 0.064  \\
    Ours & 0.3097 & \textbf{0.991} & \textbf{0.035}  \\
    \bottomrule
  \end{tabular}
  \caption{Quantitative comparisons with Rerender-A-Video (RAV), CoDeF and TokenFlow. Our method outperforms others in temporal consistency.}
  \label{tab:quantitative}
  \vspace{-5mm}
\end{table}

% \gaoxuan{Should we add the experiments on the balance between $s_I$ and $s_T$?}

% \haiyao{maybe not? it seems that this part may have a greater impact on the convergence speed rather than the final result, unless there are significant variations in the input data.}

\subsection{Ablation Study}

\begin{figure}[t]
  \centering
   \includegraphics[width=1.0\linewidth]{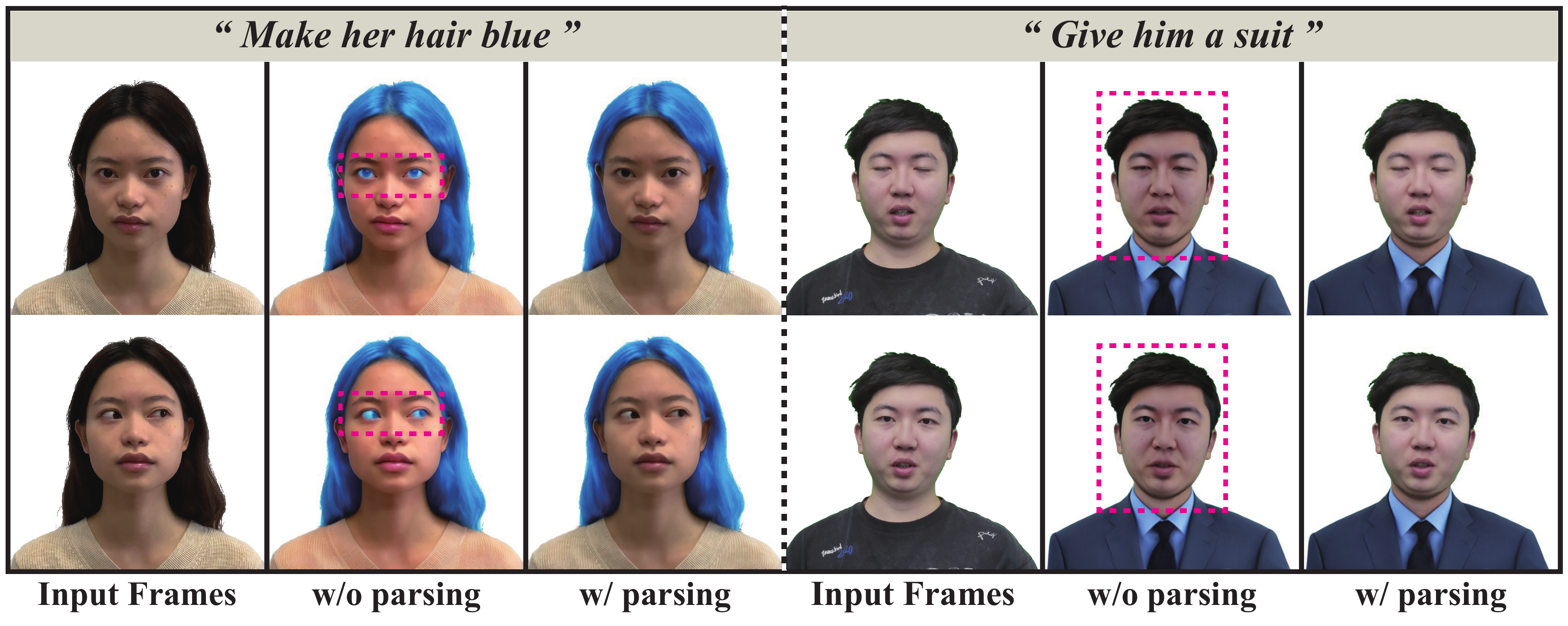}
   \caption{Ablation study on incorporating semantic segmentation. The results demonstrate that leveraging semantic priors significantly enhances the accuracy of local attribute editing.}
   \label{fig:ablation}
   \vspace*{-4mm}
\end{figure}

% As aforementioned in Sec. \ref{method32}, we use FLAME coefficients as control conditions for the NeRF model. The utilization of FLAME parameters offers a wealth of semantic information, which enhances the representation of motion in portraits, particularly in capturing facial expression variations. By integrating expression parameters as conditions within the deformation network and appearance feature, there is a notable enhancement in both the convergence speed and rendering quality during the pre-training phase of NeRF model. Furthermore, the expression parameters exhibit strong temporal consistency as they are directly derived from the original video. In Figure *, we endeavor to substitute the FLAME coefficients with optimized codes and concurrently optimize them alongside the pre-training phase with the NeRF model. Furthermore, we introduce inter-frame constraints to guarantee the temporal coherence of the optimized codes. From the illustration results, it can be clearly observed that The incorporation of semantic parameters contributes to enhanced coherence in the edited results.
% The second part of our ablation study investigates the importance of incorporating semantic segmentation.
We conduct ablation study to validate the effectiveness of incorporating semantic segmentation in our method. Although Instruct-Pix2Pix can produce editing results that align with text instructions, it might fail to accurately modify specific local areas as intended, leading to unintended and excessive alterations to the image. As illustrated in Fig.~\ref{fig:ablation}, modifying the portrait's hair or clothing may affect other areas, causing the NeRF model to acquire inaccurate information. Furthermore, these erroneous editing results will persist and have a cumulative impact during subsequent editing procedures because of the dataset update strategy. To resolve this issue, we leverage prior knowledge from semantic segmentation to meticulously modify the generated images, ensuring precise editing of local attributes.

% The rich semantic prior helps to perform more accurate local attribute editing.

% The results demonstrate that leveraging semantic priors enhances the accuracy of local attribute editing.

% \begin{table}[t]
%  \centering
%  \renewcommand{\arraystretch}{1.3}

% \begin{tabular}{p{1.5cm}|p{1.5cm} p{1.5cm} p{1.5cm} p{1.5cm}}
% \hline
% % \cline{2-7}
%     & CoDeF& TokenFlow& Rerender-A-Video & Ours \\
%   \hline CLIP-score& 4.06& 1.64& 0.166& 0.043 \\
%    \hline Tem-Con &3.962 &1.63& 0.145 &0.011  \\
%      \hline Pixel-MSE& \textbf{3.242}& \textbf{0.758}&\textbf{0.129} &\textbf{0.002}   \\
%     % \hline Source & & & 0.132& & \\
% \hline
% \end{tabular}
%  \caption{}
% \label{tab:ablation of designs of the network}
% \end{table}

\subsection{Application}

As shown in Fig.~\ref{fig:exp_total} and Fig.~\ref{fig:exp_region}, our method can handle various text-driven visual editing tasks, including global style editing and attribute editing of portrait local regions. Our method performs well in portrait editing with extreme poses and expressions, generating 3D-consistent and temporally coherent results. Please refer to the supplemental material and the accompanying video for more results.

\begin{figure}[htbp]
  \centering
   \includegraphics[width=\linewidth]{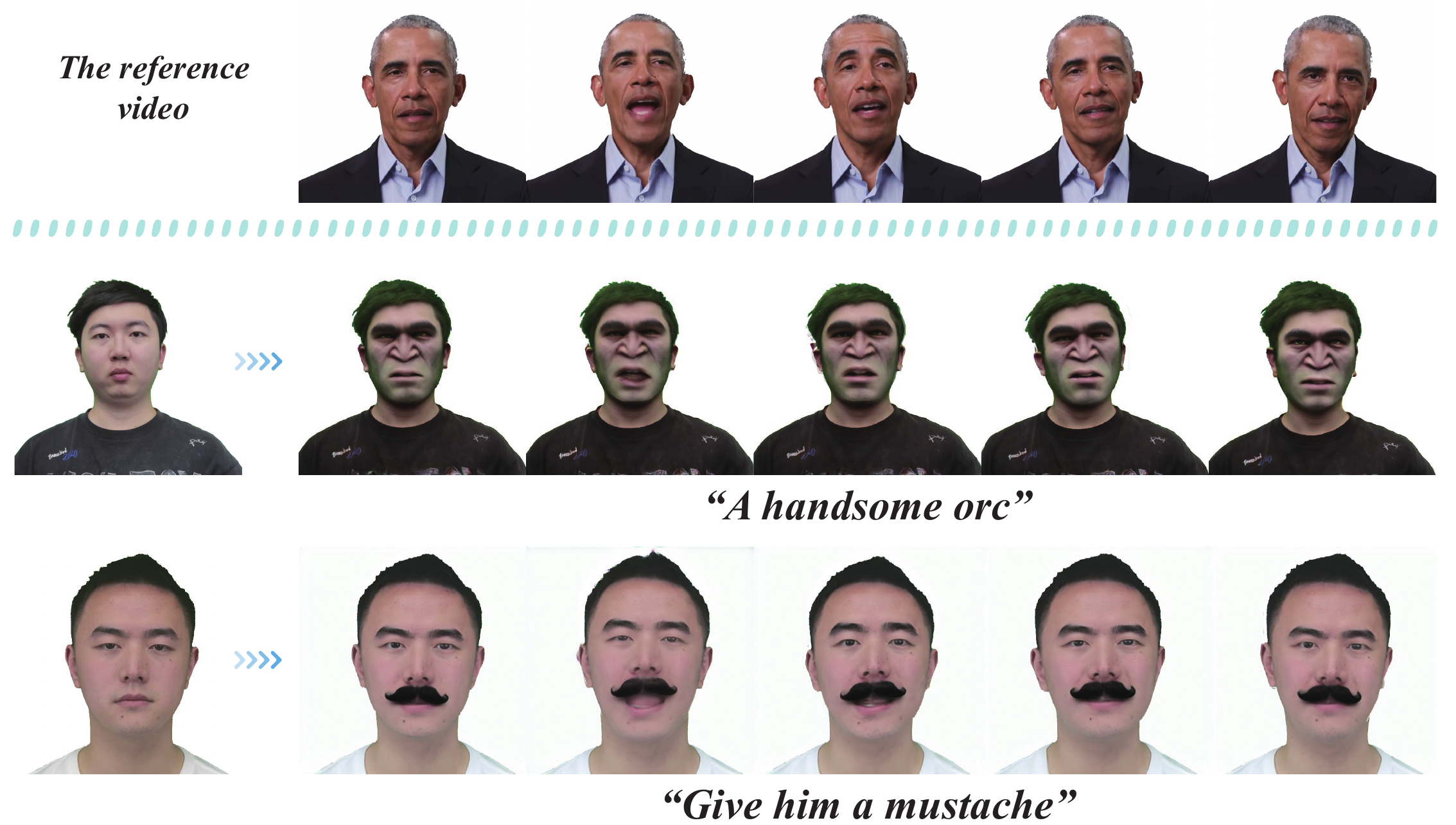}
   \caption{Expression Transfer. CosAvatar is able to transfer facial expressions from the reference video to the persons in edit results.}
   \label{fig:driving}
   \vspace*{-4mm}
\end{figure}

Additionally, by utilizing a NeRF model with FLAME coefficients as conditional inputs, our method enables the direct transfer of facial expressions and head poses from the reference video to the edited portrait. Precisely, we can extract expression coefficients and pose information from the reference portrait videos and replace the corresponding part of the input sequence with this extracted one. Finally, we employ our model to generate the desired portrait image sequence, maintaining the appearance of editing results as well as the corresponding expressions and poses from the reference video. The results are shown in Fig.~\ref{fig:driving}.

\begin{figure}[htbp]
  \centering
   \includegraphics[width=\linewidth]{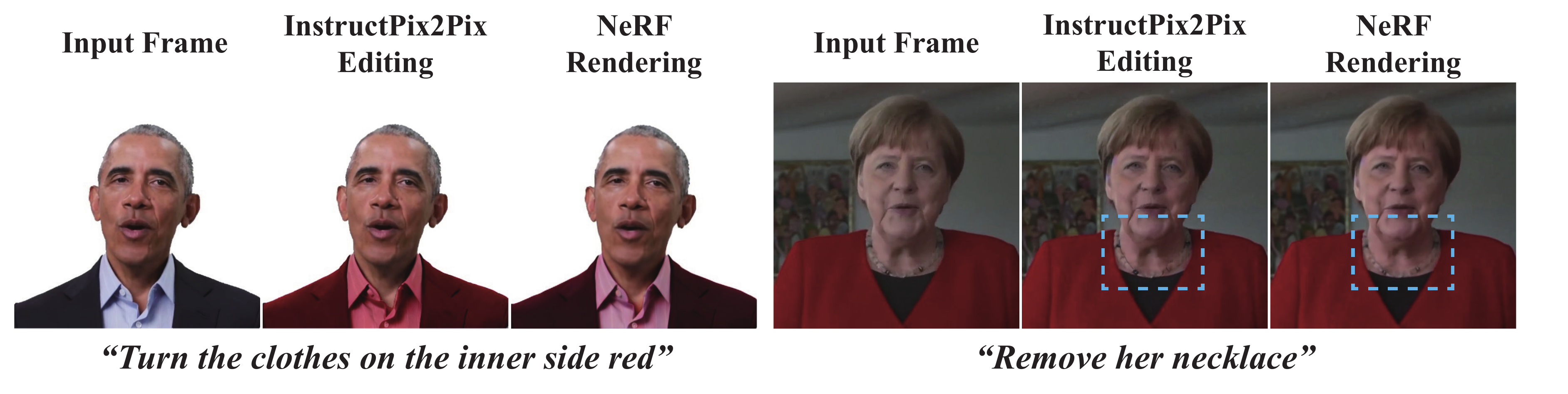}
   \caption{Limitations: InstructPix2Pix fails to isolate the specified object and spatial reasoning, thus our method is unable to achieve accurate fine-detail editing. }
   \label{fig:limitation}
    \vspace*{-4mm}
\end{figure}

\section{Conclusion and Discussion}
In this paper, we proposed CosAvatar, a high-quality and user-friendly portrait editing framework combining text-driven diffusion models with parametric NeRF head representation. Thanks to the 3D priors provided by NeRF representation and the introduction of semantic segmentation information, CosAvatar can generate consistent and animatable portrait videos and support accurate editing of local attributes. Extensive experimental results have demonstrated that the results generated by our method outperform state-of-the-art approaches in terms of temporal and 3D consistency. We believe that CosAvatar has taken a significant step towards cross-modality editing of digital humans.

\noindent{\bf{Limitations and Future Work.}} Our method still inherits some limitations from InstructPix2Pix. As shown in Fig.~\ref{fig:limitation}, although the integration of semantic segmentation contributes to more accurate editing of local attributes, our current method tends to focus on editing larger-scale areas such as hair or clothing. Because of the difficulties associated with InstructPix2Pix in isolating particular objects and spatial reasoning, our approach is unable to precisely modify finer details in the portrait.

Currently, our method lacks explicit geometric proxies for the body parts. As a result, when there are significant body movements, it may fail to capture the correct dynamic information. In the future, we plan to replace the head tracking based on FLAME~\cite{li2017learning} with half-body tracking based on SMPL-X~\cite{SMPL-X:2019} to achieve more accurate modeling of the body and capture its dynamic movements more effectively.
{
    \small
    \bibliographystyle{ieeenat_fullname}
    \bibliography{main}
}

\newpage
\setcounter{page}{1}
\maketitlesupplementary
% \appendix

% \vspace*{-9mm}

% \begin{center}
%    \begin{overpic}
%         [width=\linewidth]{figures/teaser-v4.pdf}
%         %  \fbox{\rule{0pt}{2in} \rule{linewidth}{0pt}}
%    \end{overpic}
% \end{center}
% \vspace*{-5mm}
% \captionof{figure}{CosAvatar, a text-driven portrait editing framework based on monocular dynamic NeRF. It allows for both (a) global style editing and (b) local attribute editing while ensuring strong consistency. It also enables expressive animation of the edited portrait.}

% \label{fig:teaser}

% \vspace*{5mm}

\section{Additional Details of Network Designs}
We learn our deform field $\mathcal{D}_\theta$ and conditional radiance field $\mathcal{F}_\theta$ using multilayer perceptron(MLP). In this section, we provide the implementation details of our network designs.

\noindent \textbf{Deform Field.} The deformation field takes spatial positions $\mathbf{x}$ and expression parameters $\mathbf{z}_{\textrm{exp}}$ as inputs. For spatial positions, we introduce a positional encoding layer that utilizes a Fourier positional encoding with a maximum frequency of 4. 
% \gaoxuan{For spatial positions, we use the Fourier positional encoding layer introduced in NeRF~\cite{mildenhall2020nerf} with $L=4$} 
% We use a 6-layer MLP with ReLU activation function, and each hidden layer contains 64 channels.\gaoxuan{We use a 6-layer MLP with 64 neurons width and ReLU activation function.}
\begin{figure}[h]
    \centering
   \includegraphics[width=1.0\linewidth]{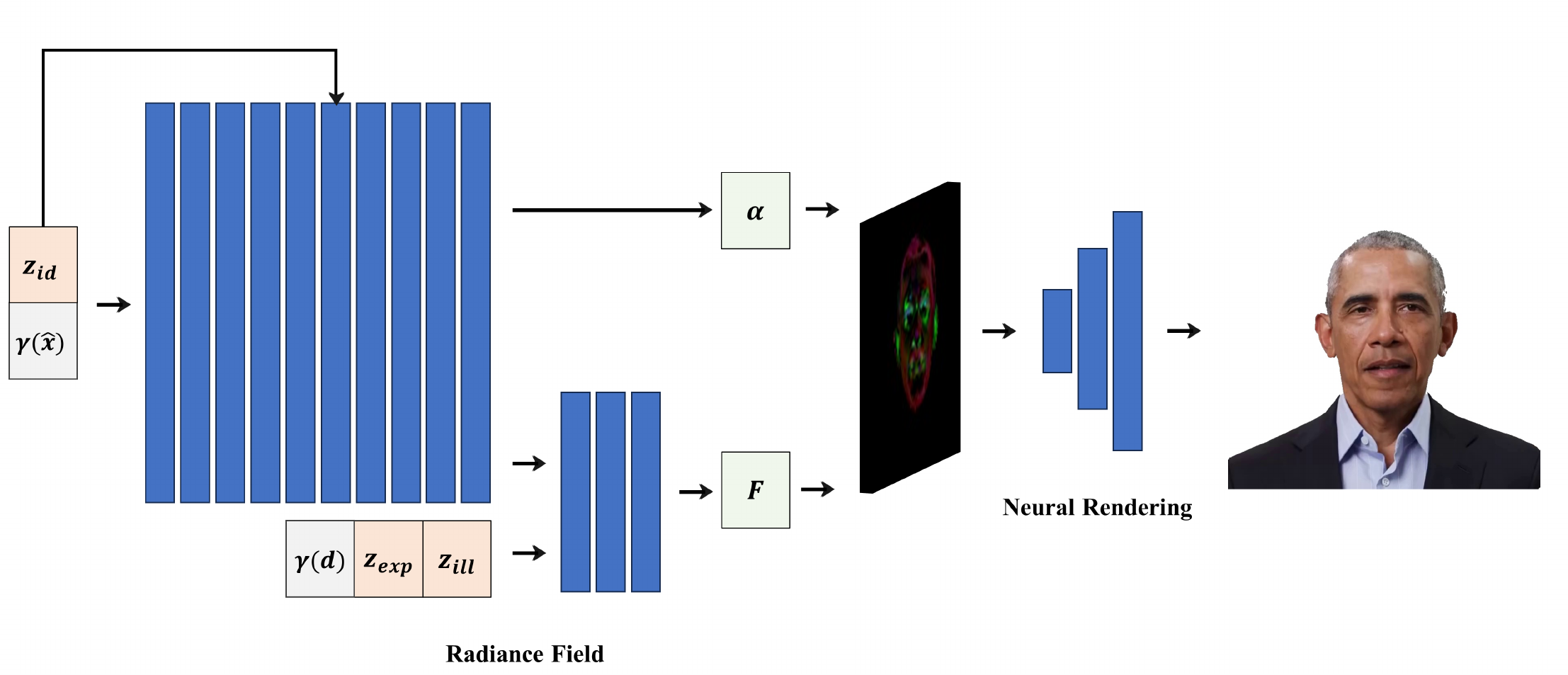}
   \caption{Network architecture of the conditional radiance field}
   \label{fig:nerf}
\end{figure}

\noindent \textbf{Conditional Radiance Field.} As shown in Fig.~\ref{fig:nerf}, the conditional radiance field first 
takes positional encoding of the input location $\gamma(\textbf{x})$ and identity prior signal $\mathbf{z}_{\textrm{id}}$ as inputs. They are passed through a 10-layer MLP with ReLU activation function, each with 512 channels. An additional layer outputs the volume density $\sigma$ and a feature vector. This feature vector is concatenated with the positional encoding of the input view direction $\gamma(\textbf{d})$ and the prior signal of expression and illumination $\mathbf{z}_{\textrm{exp}}, \mathbf{z}_{\textrm{ill}}$. Then, they are passed through a 3-layer MLP with 256 channels to output the final feature. The maximum frequency for positional encoding of input location and view direction are set to 8 and 4.
% \gaoxuan{$L$ is set separately to 8 and 4 for positional encoding of input location and view direction}

Nevertheless, employing the identity priors extracted from FLAME fails to capture the intricate shape and geometric details of real faces. To address this, we adopt a two-step approach. Firstly, we extract features using the face recognition model AdaFace \cite{kim2022adaface}. Subsequently, we pre-train an encoder to map these features to a lower dimension, 100, which serves as the input for the identity prior, thereby capturing more nuanced facial characteristics. 

For the super-resolution module, we employ the architecture directly from MetaHead~\cite{zhang2023metahead}. For specific details, please refer to their work.

\noindent \textbf{Traing Strategy.} In order to achieve efficient reconstruction of new data, we aim for our model to possess a certain level of generalization ability. To accomplish this, we commence by pretraining our model on a large-scale dataset of head images, following the aforementioned design. Subsequently, for all experiments, we only need to perform fine-tuning of the network based on the pre-trained model. This approach considerably speeds up the reconstruction process.

% \section{Settings for The Comparison Method}
% The results of Rerender-A-Video\cite{yang2023rerender}, TokenFlow\cite{tokenflow2023},
% and CoDeF\cite{ouyang2023codef} on our collected datasets are using their official code. 

\section{Implementation Details of Driving}
Our goal is to utilize the portrait from a reference video to drive the edited portrait in the original video. Initially, we employ face tracking or a pre-trained encoder to extract pose and expression information from the portrait in the reference video. The former method provides more accurate results, while the latter option has a lower computational cost.

Once we acquire the pose and expression information, we combine it with the identity information from the original video. We calculate the geometric results of the corresponding FLAME model and generate a depth map for guided sampling. At this stage, our model can directly generate the edited portrait, preserving the editing results' appearance, as well as the corresponding expressions and poses from the reference video. However, since there is no explicit geometric prior for the body during reconstruction, we cannot effectively control the body part by directly modifying the driving signal. As a solution, we utilize the joint information of the generated FLAME model's neck region to compute a set of rotations and displacements. This enables us to refine the camera for the half-body part, thereby aiming to achieve improved overall results.

\section{Discussion}
\noindent \textbf{Ethics Statement} Our digital portrait editing framework, CosAvatar, focuses on technical development. Our method can generate portrait editing results based on arbitrary textual instructions. Due to its ability to produce high-fidelity results and its high degree of flexibility in the generation process, misuse of our methods may raise ethical issues. Therefore, it is imperative to strictly prohibit any inappropriate behavior associated with its use. As a result, we require that the media data generated by our method clearly present itself as synthetic. Furthermore, we strongly believe that it is crucial to develop safeguarding measures to mitigate the potential for misuse.

% \section{Rationale}
% \label{sec:rationale}
% % 
% Having the supplementary compiled together with the main paper means that:
% % 
% \begin{itemize}
% \item The supplementary can back-reference sections of the main paper, for example, we can refer to \cref{sec:intro};
% \item The main paper can forward reference sub-sections within the supplementary explicitly (e.g. referring to a particular experiment); 
% \item When submitted to arXiv, the supplementary will already included at the end of the paper.
% \end{itemize}
% % 
% To split the supplementary pages from the main paper, you can use \href{https://support.apple.com/en-ca/guide/preview/prvw11793/mac#:~:text=Delete%20a%20page%20from%20a,or%20choose%20Edit%20%3E%20Delete).}{Preview (on macOS)}, \href{https://www.adobe.com/acrobat/how-to/delete-pages-from-pdf.html#:~:text=Choose%20%E2%80%9CTools%E2%80%9D%20%3E%20%E2%80%9COrganize,or%20pages%20from%20the%20file.}{Adobe Acrobat} (on all OSs), as well as \href{https://superuser.com/questions/517986/is-it-possible-to-delete-some-pages-of-a-pdf-document}{command line tools}.

\end{document}

% \title{\LaTeX\ Guidelines for Author Response}  % **** Enter the paper title here

% \maketitle
\thispagestyle{empty}
\appendix

%%%%%%%%% BODY TEXT - ENTER YOUR RESPONSE BELOW
\section{Introduction}